\documentclass[conference]{IEEEtran}
\IEEEoverridecommandlockouts
\usepackage{cite}

\usepackage{amsmath,amssymb,amsfonts,mathtools,bm}
\usepackage{amsthm}
\usepackage{algorithmic}
\usepackage{graphicx}
\usepackage{textcomp}
\usepackage{xcolor,float}
\usepackage{tabularx}
\usepackage{textcomp}
\usepackage{diagbox}
\usepackage{svg}
\usepackage{booktabs}
\usepackage{subfigure}
\usepackage[bookmarks=false]{hyperref}
\usepackage[ruled,lined]{algorithm2e}
\usepackage{tabularx}
\usepackage{booktabs}

\newtheorem{problem}{Problem}

\allowdisplaybreaks
\renewcommand{\baselinestretch}{1}

\begin{document}

\title{Effectively Heterogeneous Federated Learning: A Pairing and Split Learning Based Approach} 
\author{
\IEEEauthorblockN{
Jinglong Shen\IEEEauthorrefmark{1},
Xiucheng Wang\IEEEauthorrefmark{1},
Nan Cheng\IEEEauthorrefmark{1},
Longfei Ma\IEEEauthorrefmark{1},
Conghao Zhou\IEEEauthorrefmark{2},
Yuan Zhang\IEEEauthorrefmark{3},
\\
}
\IEEEauthorblockA{
\IEEEauthorrefmark{1}School of Telecommunications Engineering, Xidian University, Xi'an, China\\
\IEEEauthorrefmark{2}Department of Electrical \& Computer Engineering, University of Waterloo, Canada\\
\IEEEauthorrefmark{3}School of CSE, University of Electronic Science and Technology of China, China\\
Email: \{jlshen, xcwang\_1, lfma\}@stu.xidian.edu.cn, dr.nan.cheng@ieee.org, c89zhou@uwaterloo.ca, ZY\_LoYe@126.com
}

}
    \maketitle

\IEEEdisplaynontitleabstractindextext

\IEEEpeerreviewmaketitle

\begin{abstract}
As a promising paradigm federated Learning (FL) is widely used in privacy-preserving machine learning, which allows distributed devices to collaboratively train a model while avoiding data transmission among clients. Despite its immense potential, the FL suffers from bottlenecks in training speed due to client heterogeneity, leading to escalated training latency and straggling server aggregation. To deal with this challenge, a novel split federated learning (SFL) framework that pairs clients with different computational resources is proposed, where clients are paired based on computing resources and communication rates among clients, meanwhile the neural network model is split into two parts at the logical level, and each client only computes the part assigned to it by using the SL to achieve forward inference and backward training. Moreover, to effectively deal with the client pairing problem, a heuristic greedy algorithm is proposed by reconstructing the optimization of training latency as a graph edge selection problem. Simulation results show the proposed method can significantly improve the FL training speed and achieve high performance both in independent identical distribution (IID) and Non-IID data distribution.
\end{abstract}


\section{Introduction}
As a revolutionary solution for data privacy and distributed computational learning, federated learning (FL) has gained immense popularity in recent years, since its remarkable achievements in improving the speed of distributed training significantly, while ensuring data privacy without data transmission among clients during training \cite{li2021survey,10061663}. This is achieved by bringing together the local training efforts of each client's individual model, which is trained exclusively using their unique dataset. In the traditional practice\cite{pmlr-v54-mcmahan17a} of each communication round, the server distributes a global model with aggregated parameters, and each client uses their private dataset to train the model locally. Afterward, the local models are transmitted to the server for aggregation. Once all the local models are accumulated, the server aggregates them, shares the aggregated model with the clients, and begins a new training epoch \cite{shen2021joint}.

However, the heterogeneity of clients imposes on the performance of FL systems, which leads to an escalation in training latency due to the limited computing resources of certain clients that could straggle the server’s aggregation process \cite{li2021survey}. Although, the widely-used edge computing (EC) technology can be used to alleviate the phenomena of stragglers by providing the computing resources of the edge server for clients \cite{wang2022digital}. However, in traditional EC, clients generally need to offload their data to the server, which violates data privacy protection in FL. Fortunately, the forward inference and the backward training of the model are sequential, which enables the neural network (NN) to be split into a few parts and processed in serial \cite{Shen2023}. Therefore, in \cite{tian2022fedbert} the NN is split into three parts, each client has the up and bottom parts whose size is small, and the large middle part is offloaded to the server to process. However, when there are lots of clients, the computing burden of the server is too heavy to alleviate the phenomena of stragglers effectively, moreover, offloading middle parts to the server can be time-consuming for clients who are far away from the server. Thus, designing a novel approach that can effectively tackle the straggler issues in both computing and communication latency remains a crucial area of research in optimizing the training speed of FL.

Inspired by the pervasive EC, which allows clients with limited computational resources to offload to other clients with abundant resources \cite{9197692}, a client-pairing-based split federated learning (SFL) framework, named FedPairing, is proposed in this paper, to deal with the challenge of stragglers. In this framework, clients with disparate computational resources are paired while considering the communication rates between paired clients, enabling each client to divide its local model into two parts based on computational resources of itself and its paired client, and each client only computes part of the local model assigned to it. Remarkably, the upper part containing the input layer is processed by the client itself, thus avoiding data transmission between paired clients and safeguarding data privacy. Under this framework, each client has two sources of gradients during the backward procedure: the gradient transmitted from the paired client, owing to split learning, and the gradient computed by the client itself. This structure enables clients to indirectly train their local models with a larger dataset, which is helpful to improve the training accuracy of the FL. To pair clients efficiently, a heuristic greedy-based algorithm is proposed, which takes into account the transmission rates among clients and the computing resources of each client, thereby enhancing the training speed. The main contributions of this paper are as follows.
\begin{enumerate}
    \item A novel SFL framework, named FedPairing, is proposed that pairs clients with different computing resources, aiding in managing the straggler phenomenon and enhancing data privacy within Federated Learning.
    \item A heuristic greedy algorithm is proposed to optimize client pairing by reconstructing the optimization of training latency as a graph edge selection problem, which increases the effectiveness of the Federated Learning system.
    \item  Simulation results show the proposed method can significantly improve the FL training speed and achieve high performance both in independent identical distribution (IID) and Non-IID data distribution. 
\end{enumerate}

\section{Client Pairing-based SFL Framework}
\begin{figure*}[ht]
    \centering
    \includegraphics[width=1.6\columnwidth]{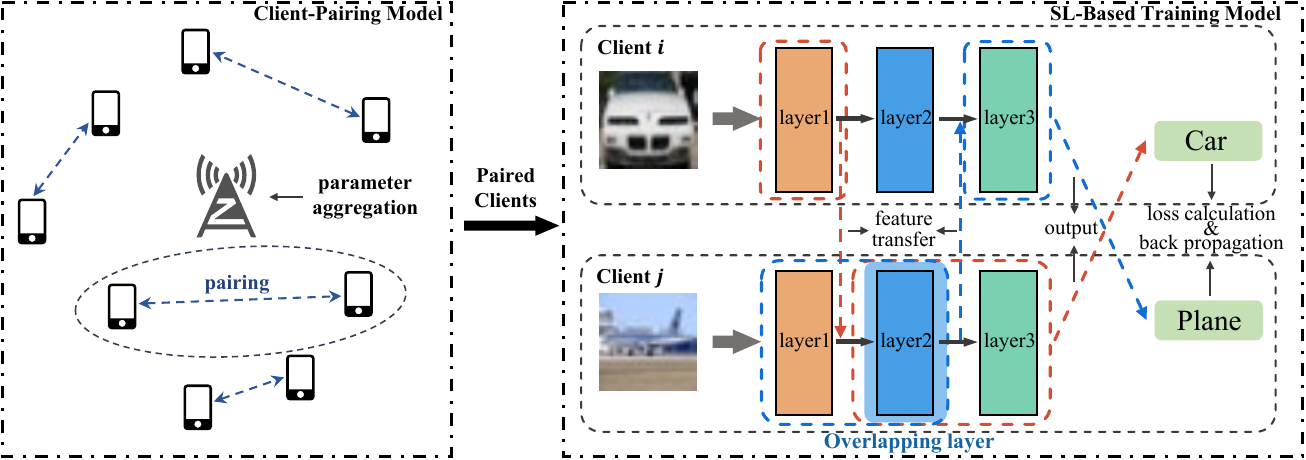}
    \caption{The architecture of FedPairing.}
    \label{fig:model}
\end{figure*}
\subsection{FedPairing Training Process}
As shown in Fig.~\ref{fig:model}, our approach considers a scenario comprising $N$ clients and a parameter aggregation server. Each client $c_i$ holds a training dataset $\mathcal{D}_i$ of size $|\mathcal{D}_i|$ and operates at different CPU frequencies, $f_i$. Clients can upload local model parameters $\omega^i$ with $W$ layers to the server and can also establish communication connections with other clients, controlled by the server, to implement logical-level model splitting for local training.

For a mini-batch $(\bm{x}_i, y_i)$ uniformly sampled from $\mathcal{D}_i$, where $\bm{x}_i$ represents the training sample and $y_i$ denotes the corresponding label, the forward propagation of the model $\omega^{i}$ is expressed as $\hat{y}_i=\omega^{i}(\bm{x}_i)=\omega^{i}_{(1,W)}(\bm{x}_i)=\omega^{i}_{(W)}(\omega^{i}_{(W-1)}(\cdots \omega^{i}_{(1)}(\bm{x}_i)))$. Here, $\omega^{i}_{(1,W)}$ represents all layers from $1$ to $W$ in $\omega^i$, and $\omega^{i}_{(j)}$ denotes the $j$-th layer of $\omega^i$. Similarly, the backward propagation is expressed as $\tilde{\omega}^{i}(loss)=\tilde{\omega}^{i}_{(1,W)}(loss)=\tilde{\omega}^{i}_{(1)}( \tilde{\omega}^{i}_{(2)}( (\cdots \tilde{\omega}^{i}_{(W)}(loss) ) )$. During training, the server pairs clients and performs the logical-level split training between each paired client. The FedPairing algorithm repeats this process until the global model converges.

\subsubsection{Initialization}
At the beginning of the training process, each client sends its state information $(f_i, |\mathcal{D}_i|)$ to the server. Based on this information, the server pairs clients, obtaining a set of pairs $\mathcal{E}=\lbrace \cdots, (c_i, c_j), \cdots \rbrace$. Next, the server computes the propagation lengths for each pair $(c_i, c_j)$, denoting them by $L_i$ and $L_j$, as $\left\lfloor \cfrac{f_i}{f_i + f_j}W \right\rfloor$ and $W - L_i$, respectively. Moreover, the server determines the aggregation weight $a_i$ for each client based on the same method used in FedAvg\cite{pmlr-v54-mcmahan17a}. Specifically, $a_i$ is computed as $\cfrac{|\mathcal{D}_i|}{\sum_{j=1}^N|\mathcal{D}_j|}$. In the final step, each client establishes a communication connection with its paired client, and the server distributes the initial global model $\omega^g$, along with respective $(L_i, a_i)$ values, to the corresponding clients.

\subsubsection{Local Training Process}
To address the straggler effect in our training process, a logic-level model splitting mechanism is employed for local training. This mechanism assigns smaller propagation lengths to clients with poorer computational power in each pair, thus reducing their training load. Specifically, for each pairs $(c_i,c_j)$, both forward and backward propagation processes of $c_i$ and $c_j$ are executed identically and in parallel. During each communication round, multiple forwards and backward propagation are performed. For brevity, the rest of this paper will use the forward and backward propagation process of $c_i$ as an example.

\textbf{Forward propagation}: During training, a client $c_i$ samples a mini-batch $(\bm{x}_i, y_i)$ from its local dataset $\mathcal{D}_i$. It then feeds $\bm{x}_i$ into its local model $\omega_i$ to forward propagate $L_i$ layers, generating the corresponding feature map, $\bar{\bm{x}}_i=\omega^{i}_{(1,L_i)}(\bm{x}_i)$. Next, the client sends $\bar{\bm{x}}_i$ to $c_j$, which further propagates the feature map through the remaining layers, $L_j=W-L_i$, in its local model to compute the output, $\hat{y}_i$, i.e., $\hat{y}_i=\omega^{j}_{(L_i+1,W)}(\bar{\bm{x}}_i)$. Finally, $c_j$ returns $\hat{y}_i$ to $c_i$, and the loss value is calculated as $l_i=CrossEntropy(\hat{y},y)$. We also note that the forward propagation process of $c_i$ can be represented as $\omega^{j}_{(L_i+1,W)}(\omega^{i}_{(1,L_i)}(\bm{x}_i)),$ and that of $c_j$ can be formulated as $\omega^{i}_{(L_j+1,W)}(\omega^{j}_{(1,L_j)}(\bm{x}_j))$.

\textbf{Backward propagation}: The client $c_i$ sends the loss value $l_i$ and its own aggregation weight $a_i$ to $c_j$. Subsequently, $c_j$ back-propagates $L_j=W-L_i$ layers on the local model $\omega^j$ to obtain the gradients $\bm{g}^{i}_{(L_i+1,W)}$ for the layers from $L_i+1$ to $W$. These gradients are then weighted by $a_i$ and cached locally. $c_j$ then sends the gradient $\bm{g}^{i}_{(L_i+1)} = \tilde{\omega}^{j}_{(L_i+1,W)}(l_i)$ of the $L_i+1$-th layer to $c_i$. In the backward propagation process, $c_i$ back-propagates $L_i$ layers in its local model to obtain the gradient $\bm{g}^{i}_{(1,L_i)}$ for the layers from $1$ to $L_i$, which is also weighted by $a_i$ and cached locally. The backward propagation process of $c_i$ is denoted by $\tilde{\omega}^{i}_{(1,L_i)}(\tilde{\omega}^{j}_{(L_i+1,W)}(l_i))$, and that of $c_j$ is denoted by $\tilde{\omega}^{j}_{(1,L_j)}(\tilde{\omega}^{i}_{(L_j+1,W)}(l_i))$.

When both clients $c_i$ and $c_j$ complete backward propagation, they update their local model using cached gradients, which can be formulated as
\begin{align}
    \omega^i = \omega^i - \eta \left(a_i\bm{g}^{i}_{(1\,,\,L_i)} + a_j\bm{g}^{j}_{(W-L_i\,,\,W)}\right), \\
    \omega^j = \omega^j - \eta \left( a_j\bm{g}^{j}_{(1\,,\,L_j)} + a_i\bm{g}^{i}_{(W-L_j\,,\,W)} \right).
\end{align}

Since the propagation length constrains the number of layers propagated locally, we can control the training load of $c_i$ and $c_j$ by adjusting the propagation lengths.
Specifically, to avoid either side in each pair to become straggler, in this paper, we adjust the propagation length of $c_i$ and $c_j$ so that the time taken for $c_i$ and $c_j$ to complete forward and backward propagation is as equal as possible. We denote the average number of CPU cycles required to update a neural layer once (including forward propagation, backward propagation and parameter update) as $F$. The propagation time on $\omega^i$ is $\cfrac{L_iF}{f_i}$, and the propagation time on $\omega^j$ is $\cfrac{L_jF}{f_j}$. If the propagation time consumed on $\omega^i$ and $\omega^j$ is equal, then we have $\cfrac{L_iF}{f_i}=\cfrac{L_jF}{f_j}\rightarrow\cfrac{L_i}{f_i}=\cfrac{L_j}{f_j}$. This means that when the ratio of propagation length and the ratio of CPU frequency are equal, both propagation time consumption are equal, so we set the propagation length of $c_i$ as $L_i=\left\lfloor \cfrac{f_i}{f_i+f_j}W \right\rfloor$, and the propagation length of $c_j$ as $L_j=W-L_i$.

During the local training process, it is obvious that paired clients transmit many intermediate results, such as forward feature maps and the backward gradient. As a result, transmission latency cannot be ignored. Furthermore, to emphasize the importance of client pairing, this paper does not take interference into consideration, and the orthogonal frequency division multiplexing (OFDM) technique is used to establish communication connections among clients.
Then the communication rate between $c_i$ and $c_j$ is
\begin{equation}
    \begin{array}{cc}
        r_{i,j}=B\log_2(1+\cfrac{Ph_{i,j}}{\sigma^2}), \\
        h_{i,j}=h_0\left( \cfrac{\zeta_0}{\|\bm{p_i}-\bm{p_j}\|} \right)^\theta,
    \end{array}
\end{equation}
where $B$,$P$,$\sigma^2$ denote the spectral bandwidth, transmission power and noise power, respectively. The channel gain between clients $c_i$ and $c_j$ is determined by the distance $\|\bm{p_i}-\bm{p_j}\|$ between them, where $\bm{p_i}$, $\bm{p_j}$ denote the positions of $c_i$, $c_j$, respectively. Additionally, $h_0$ represents the reference channel gain for a unit distance $\zeta_0$, while $\theta$ signifies the path-loss exponent.

\subsubsection{Model Aggregation}

In each communication round, each client needs to upload its local model to the server for aggregation to achieve model consensus after completing local training. Since the aggregation weights of each client have already been applied to the corresponding gradients during backward propagation, the server can directly perform averaging to complete the model aggregation $\omega^{g}=\sum_{i=1}^N\omega^{i}$. FedPairing will perform multiple communication rounds until the global model converges.

\subsection{Problem Formulation}
In order to enhance the training efficiency that is affected by both computing and communication delays, we formulate the problem as the follow
\begin{problem}\label{p1}
    \begin{align}
        &\min_{\bm{\kappa}} \sum_{i=1}^{N}\sum_{j=1}^{N} \alpha\frac{L_iF}{f_i}  \notag\\
        &+\beta\kappa_{(c_i,c_j)}\frac{\max\{|\mathcal{D}_i|\bar{x}_{i}+|\mathcal{D}_j|\bm{g}^j_{(L_j)}, |\mathcal{D}_j|\bar{x}_{j}+|\mathcal{D}_i|\bm{g}^i_{(L_i)}\}}{r_{i,j}}\label{obj}\\
        \text{s.t.} &\quad \sum_{j=1}^{i} \kappa_{(c_i,c_j)} \leq 1 \quad\;\; \forall i \in \{1,\cdots,N\},\tag{\ref{obj}{a}}\label{c-1}\\
        &\quad \kappa_{(c_i,c_j)} \in \{0,1\}, \quad \forall i,j \in \{1,\cdots,N\},\tag{\ref{obj}{b}}\label{c-2}
    \end{align}
\end{problem}
\noindent where $\alpha$ and $\beta$ are weighting constants, and $\kappa_{(c_i,c_j)}$ is a binary variable, it is equal to 1 if and only if the clients $c_i$ and $c_j$ are paired, otherwise it is 0. The objective \eqref{obj} is to minimize the computing delay $\frac{L_iF}{f_i}$ and the intermediate results transmission latency between paired clients. \eqref{c-1} and \eqref{c-2} constraint each client can only be paired with one another client.

\section{Greedy Strategy-based Pairing Algorithm}
\subsection{Graph Modeling Based Greedy Algorithm}
Since the problem \ref{p1} is an integer linear programming problem that has been shown to be NP-hard. To be able to derive the solution in polynomial time, we propose a heuristic algorithm based on a greedy strategy by reconstructing the problem \ref{p1} as a graph edge selection problem. Firstly, clients participating in the training are modeled as a weighted undirected graph $\mathcal{G}=(\mathcal{E},\mathcal{V})$, where each vertex in the vertex set $\mathcal{V}$ denotes each client and the edges in the edge set $\mathcal{E}$ denote the corresponding two clients $(c_i,c_j )$ can establish a communication link. The pairing process of clients is accomplished by selecting a subset $\mathcal{E}^\prime$ of edges without shared vertices from the graph $\mathcal{G}$ such that the sum of the edge weights in the subset is maximized. Specifically, in order to make the total computational power of each pair of clients more balanced, the clients with the most computational power need to be paired together with the clients with the poorest computational power. In addition, clients with large inter-communication rates need to be paired together to avoid too low communication rates slowing down the training. Therefore, we define the weights of the edges between $c_i$ and $c_j$ as
\begin{equation}
    \epsilon_{i,j}=\alpha\cdot(f_i-f_j)^2 + \beta\cdot r_{i,j},
\end{equation}
where $f_i$, $f_j$ are the computational frequencies of $c_i$, $c_j$, respectively, $r_{i,j}$ denotes the communication rate between $c_i$ and $c_j$. Then the edge selection problem can be formulated as follows.
\begin{problem}\label{p2}
\begin{align}
\max_{\bm{\kappa} \in \{0,1\}^{|\mathcal{E}|}} &\quad \sum_{(c_i,c_j) \in \mathcal{E}} \epsilon_{i,j}\kappa_{(c_i,c_j)}\label{equa:match_obj} \\
\text{s.t.} &\quad \sum_{(c_i,c_j) \in \mathcal{E}_{c_i}} \kappa_{(c_i,c_j)} \leq 1 \quad \forall c_i \in \mathcal{V},\tag{\ref{equa:match_obj}{a}}\label{equa:constrainA}\\
&\quad \kappa_{(c_i,c_j)} \in \{0,1\}, \quad \forall (c_i,c_j) \in \mathcal{E},\tag{\ref{equa:match_obj}{b}}\label{equa:constrainB}
\end{align}
\end{problem} 
\noindent where \eqref{equa:constrainA} and \eqref{equa:constrainB} are equal to \eqref{c-1} and \eqref{c-2}. $\mathcal{E}_{c_i}$ denotes the set of all edges with $c_i$ as one endpoint. In order to solve problem \ref{p2} efficiently, the greedy algorithm is used: 
1) first sort all edges in the graph in descending order of weight; 2) then initialize an empty set $\mathcal{E}^\prime$ to store the selected edges; 3) iteratively pick the edge with the largest weight from the sorted set of edges; 4) if neither vertex of the picked edge is covered, add this edge to $\mathcal{E}^\prime$; 5) Repeat step 3) until all edges are traversed.
The details of the method is described in Algorithm \ref{alg:pairing_process}. This algorithm ensures that the selected edges cover all vertices, have no common vertices, and the sum of weights is maximized. This is because we always prefer to select edges with larger weights and only add them to the selected edge set $\mathcal{E}^\prime$ if both vertices of the edge are not yet covered. This process guarantees that the selected edges do not share any vertices, while also ensuring that the sum of their weights is maximal.

\subsection{Effect of Overlapping Layers}
\addtolength{\topmargin}{0.02in}
As previously discussed, the variability of computational resources across clients leads to differences in propagation lengths, resulting in the occurrence of overlapping layers. In this context, we begin by providing a precise definition of overlapping layer and proceed to discuss its properties.

\textbf{Overlapping Layer}: An overlapping layer is defined as a neural layer that is simultaneously impacted by the propagation flows of two clients from a given pair.


As depicted in Fig.~\ref{fig:model}, we consider a multilayer perceptron (MLP) with 3 fully connected layers that is trained using FedParing. Given any pair of clients $(c_i,c_j)$, suppose their propagation lengths are set to $L_i=1$ and $L_j=2$, respectively. In such scenario, the propagation flow of $c_i$ (indicated by red arrows in the figure) propagates through $\omega^{i}_{(1)}$ and $\omega^{j}_{(2,3)}$, while the propagation flow of $c_j$ (indicated by blue arrows in the figure) propagates through $\omega^{j}_{(1,2)}$ and $\omega^{i}_{(3)}$. Notably, in the local model of $c_j$, $\omega^{j}_{(2)}$ is present in both $\omega^{j}_{(2,3)}$ and $\omega^{j}_{(1,2)}$. This implies that the propagation flows of clients $c_i$ and $c_j$ pass through $\omega^{j}_{(2)}$, which is therefore referred to as the overlapping fully connected layer. During backward propagation, $\omega^{j}_{(2)}$ accumulates the gradients from both $c_i$ and $c_j$, thereby resulting in the aggregation of the gradients of the two clients to the second fully connected layer after each backward propagation. Since overlapping fully connected layers are subject to a higher frequency of aggregation as compared to other layers, their gradients serve as more potent drivers in bringing the parameters of overlapping fully connected layers to the global optimum, leading to an enhancement in the performance of the global model.

\begin{algorithm}[h]
    \SetAlgoLined
    \DontPrintSemicolon
    \KwIn{The graph of clients $\mathcal{G}=(\mathcal{E}, \mathcal{V})$.}
    \KwOut{Selected edges $\mathcal{E}^\prime$.}
    \BlankLine

    Sort all edges $\mathcal{E}$ in ascending order by edge weight.\;

    Initialize selected edges $\mathcal{E}^\prime$ as an empty set.\;

    Initialize covered vertexes $\mathcal{V}^\prime$ as an empty set.\;
    
    \For{edge $(u,v)$ in $\mathcal{E}$}{
        
        \If{$u$ not in $\mathcal{V}^\prime$ and $v$ not in $\mathcal{V}^\prime$}{
            
            Add $e$ into $\mathcal{E}^\prime$.\;

            Add $u, v$ into $\mathcal{V}^\prime$.\;
        
        }
    
    }
    
    \Return{$\mathcal{E}^\prime$}\;
    
    \caption{The Greedy Strategy Based Pairing Algorithm}
    \label{alg:pairing_process}
\end{algorithm}

\begin{algorithm}[h]
    \SetAlgoLined
    \DontPrintSemicolon
    \BlankLine

    The server pairs all participating clients using Algorithm~\ref{alg:pairing_process}.
    
    The server configures $L_i$ and $a_i$ for each client.
    
    The server distributes $(L_i, a_i)$ to each clients along with initialized global model $\omega_g$.

    \For{each round from 1 to $T$}{

        \For{each pair $e=(c_i, c_j)$ in pair set $\mathcal{E}$ parallelly}{
    
            \For{each client $c_i$ in $e$ parallelly}{
    
                Set $\omega_i = \omega_g$.
    
                \For{each epoch from 1 to $E$}{
    
                    \For{each batch $(\bm{x}_i, y)$ in local dataset $\mathcal{D}_i$}{
    
    
    
                        $\hat{y} = \omega_{j,(L_i+1,W)}(\omega_{i,(1,L_i)}(\bm{x}_i))$.
                        
                        $l_i = CorssEntropy(\hat{y}, y)$.
    
    
    
                        $\tilde{\omega}_{i, (1,L_i)}(\tilde{\omega}_{j, (L_i+1,W)}(l_i))$.

                        Update the parameters of $\omega_i$ based on the gradient locally cached.
                    }
                    
                }
    
                Upload trained $\omega_i$ to server.
                
            }
        
        }
    
        The server aggregates models: $\omega_g = \cfrac{1}{N}\sum_{i=1}^{N}\omega_i$.
        
        The server distributes aggregated model $\omega_g$ to clients.

    }
    
    \caption{FedPairing Training Process}
    \label{alg:training_process}
\end{algorithm}

To make better use of this property of overlapping layers, we increase the step size of the overlapping layer when updating parameters. Specifically, for any pair of clients $(c_i,c_j)$, if the $k$-th layer of $c_j$ is an overlapping layer, the iterative parameter updating process for that layer can be formulated as

\begin{align}
\omega^{i}_{(k)}=\omega^{i}_{(k)}-2\eta(a_i\bm{g}^{i}_{(k)} + a_j\bm{g}^{j}_{(k)}).
\end{align}

By enhancing the update approach adopted for the overlapping layer, we discovered that the overall performance of the global model was significantly improved. Further details on this observation are elaborated on in the simulation section.

\section{Simulation Results}

\subsection{Simulation Setup}

In the simulation, we consider the collaborative training of 20 clients distributed randomly in a 50m radius circular area. The server used to aggregate model parameters is located at the center of the area. Each client communicates using a bandwidth of 64 MHz and a transmit power of 1 W, with a noise power of 1e-9 W. The computational frequencies follow a uniform distribution from 0.1 GHZ to 2 GHZ. We execute 100 rounds of communication where each client trains locally for 2 epochs in each round.

We train the ResNet18\cite{Resnet} neural network model using stochastic gradient descent optimization with a learning rate of 0.1. The model is trained on CIFAR10\cite{cifar10-krizhevsky2009learning}, containing 50,000 samples for training and 10,000 samples for testing. The samples are divided into 10 classes with 5000 color images of 3x32x32 in each class. We evaluate the impact of dataset distribution on FedPairing using both IID and Non-IID datasets. In IID, each client's local dataset contains 2500 samples with an identical number of samples in each category. On the other hand, in Non-IID, samples containing two randomly selected categories are included in each client's local dataset. We implement the simulation in python 3.9.13 with the neural network model built and trained using PyTorch 1.11.0. For comparison purposes, we benchmarked FedPairing against three other approachs: vanilla FL\cite{pmlr-v54-mcmahan17a}, vanilla SL\cite{GUPTA20181}, and SplitFed\cite{Thapa2020}.

\subsection{Convergence Performance}

\begin{figure}[h]
    \centering
    \includegraphics[width=0.98\columnwidth]{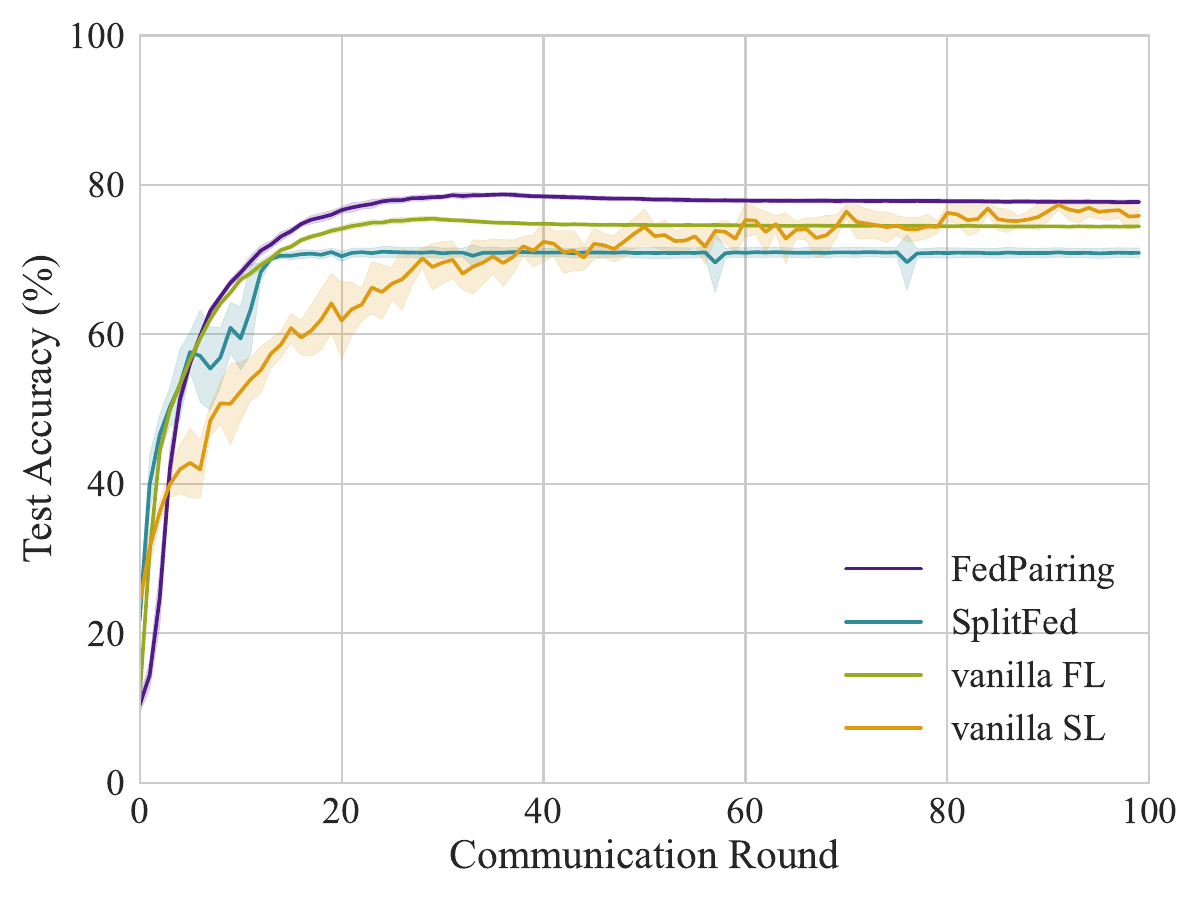}
    \caption{Convergence performance of different algorithms on IID dataset.}
    \label{fig:acc_iid}
\end{figure}

\begin{figure}[h]
    \centering
    \includegraphics[width=0.98\columnwidth]{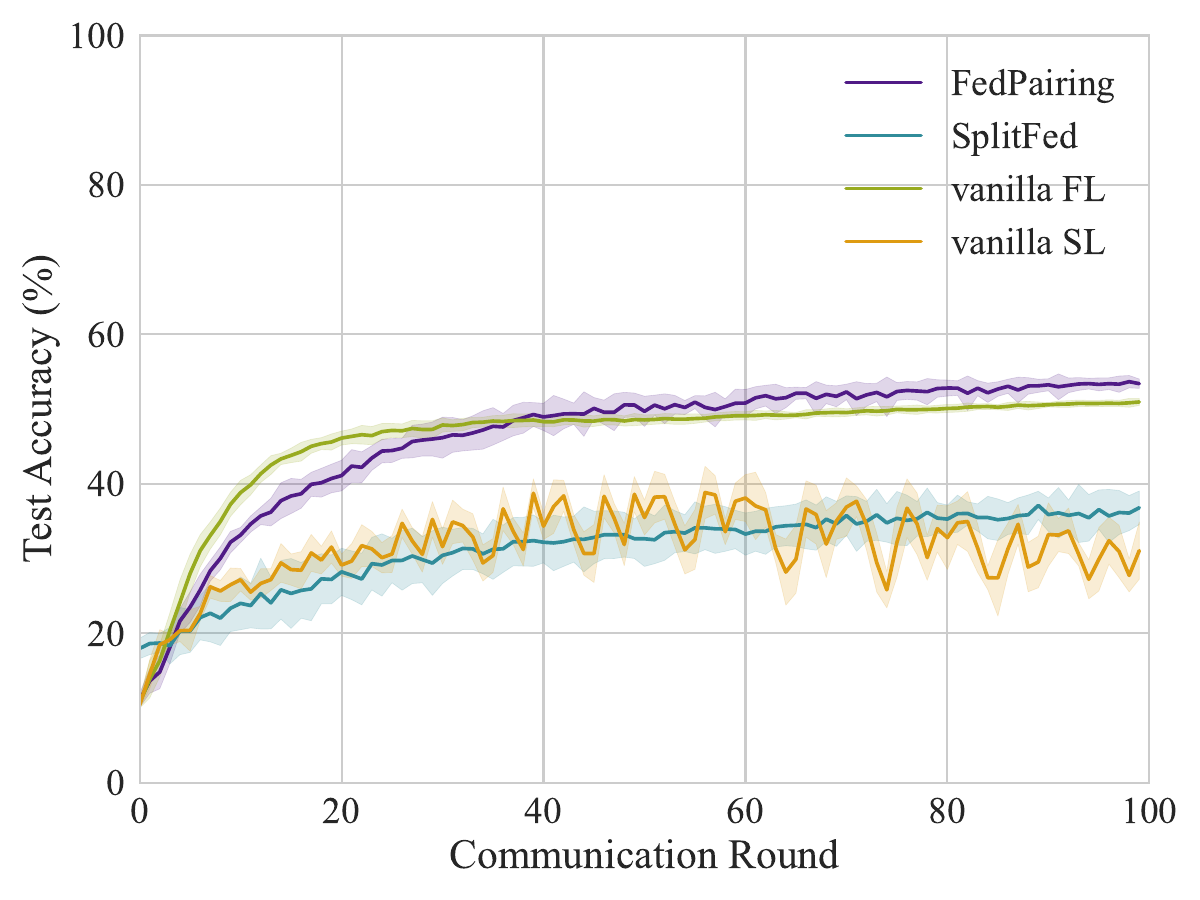}
    \caption{Convergence performance of different algorithms on Non-IID dataset.}
    \vspace{-15pt}
    \label{fig:acc_noniid}
\end{figure}

We evaluate the convergence performance of FedPairing by comparing the Top-1 accuracy of the model. The benchmarks used for comparison in the experiments are vanilla FL, vanilla SL, SplitFed. The model trained is ResNet10, and the dataset used for training is CIFAR10.

As shown in Fig.~\ref{fig:acc_iid}, FedPairing achieves the highest model accuracy at the end of training, with 4.1\%, 1.8\%, and 10.8\% improvement compared to vanilla FL, vanilla SL, and SplitFed, respectively. This is mainly due to the fact that when the propagation lengths of the two clients of a pair are not equally assigned, it leads to the existence of overlapping layers on the client with the larger propagation length. Since the gradients of these two clients on the overlapping layer accumulate with each backpropagation (meaning that the gradients of the overlapping layer have a higher aggregation frequency), it makes the gradients of the overlapping layer more capable of driving the corresponding parameters to move toward the global optimum, thus allowing FedPairing to achieve a higher model performance.

Fig.~\ref{fig:acc_noniid} shows the effect of data distributions on the FedPairinig convergence performance when trained on the Non-IID CIFAR10 dataset. It can be seen that FedPairing still maintains the best model accuracy on the Non-IID dataset. Compared to vanilla FL, vanilla SL, SplitFed, the improvement is 5.3\%, 38.2\%, and 44.6\%, respectively. This indicates that FedPairing can adapt well to the Non-IID data distribution.

\subsection{Effect of Pairing Mechanism}

\begin{table}[htbp]
    \centering
    \caption{Average Time Cost of a Communication Round under Different Pairing Mechanisms.}
    \resizebox{\columnwidth}{!}{
    \begin{tabular}{c c c c}
        \toprule
        \textbf{FedPairing} & \textbf{Random} & \textbf{Location-based} & \textbf{Computation resource-based} \\
        \midrule
        1553s & 4063s & 7275s & 1807s \\
        \bottomrule
    \end{tabular}
    }
    \label{table:pairing}
\end{table}

To evaluate the impact of pairing mechanisms on training time, we compared different pairing mechanisms of FedPairing. Specifically, we compared the following three pairing mechanisms: random pairing, location-based pairing (by calculating the geographical distance), and computation resource-based pairing (by calculating the differences in computing capabilities). By comparing these different pairing mechanisms, we expect to analyze their strengths and weaknesses in-depth and study their impact on training time.

The results presented in Table~\ref{table:pairing} show that the greedy pairing strategy of FedPairing yields the shortest training time. In comparison to random pairing, location-based pairing, and computation resource-based pairing strategies, it reduces the time consumption by 61.8\%, 78.7\%, and 14.1\%, respectively. This is because training time can be divided into communication and computation time, and the baseline strategies optimize these times via different means. Specifically, the location-based strategy only optimizes communication time, whereas the computation resource-based strategy only optimizes computation time, and joint optimization cannot be achieved. On the other hand, FedPairing's greedy pairing strategy considers both communication rate and computing resources between clients, resulting in the shortest training time. Moreover, it serves as a highly effective solution to mitigate the straggler effect.

\subsection{Training Time Cost}


\begin{table}[htbp]
    \centering
    \caption{Average Time Cost of a Communication Round under Different Algorithms.}
    \resizebox{\columnwidth}{!}{
    \begin{tabular}{c c c c}
        \toprule
        \textbf{FedPairing} & \textbf{SplitFed} & \textbf{vanilla FL} & \textbf{vanilla SL} \\
        \midrule
        1553s & 1798s & 8716s & 106s \\
        \bottomrule
    \end{tabular}
    }
    \label{table:time_cost}
\end{table}

Because of the heterogeneity among clients, conventional FL suffers from the straggler effect and the training efficiency is severely dragged down. This experiment evaluates the ability of FedPairing to mitigate the straggler effect by measuring the time consumed to complete a communication round.

As shown in Table~\ref{table:time_cost}, FedPairing reduces the time consumed to complete a communication round by 82.2\%, 13.6\% compared to vanilla FL and SplitFed, respectively. This is because FedParing achieves a equalization of the total computational power of each pair by matching two clients with large computational power differences together. The client with poor computational power can offload its training load to the paired client with strong computational power through the model splitting mechanism, thus reducing its own training time consumption, and the straggler effect suffered by the whole system is thus alleviated. However, FedPairing consumes more time to complete a communication round compared to vanilla SL. This is because vanilla SL offloads most of the training load to the server with super computing power by splitting the model between the client and the server, and the client only needs to complete a small amount of computation, so vanilla SL takes less time to complete a communication round. However, vanilla SL often suffers from convergence problems on Non-IID datasets and cannot achieve comparable performance to FedPairing.

\section{Conclusion}
In this paper, a client pairing-based SFL framework has been proposed to effectively deal with the challenge of stragglers caused by the heterogeneity of client computing resources. Since the client pairing problem is NP-Hard, a heuristic greedy method, based on reconstructing the optimization of training latency as a graph edge selection problem, has been proposed to solve the problem efficiently. Simulation results show the proposed can significantly reduce the training delay for clients with various computing resources, and achieve high performance both in IID and Non-IID datasets. By implementing this scheme in FL, the training efficiency can be significantly improved. For future research, we will study how to combine clients into groups with the arbitrary number of clients, according to their features.

\bibliography{ref}
\bibliographystyle{IEEEtran}

\ifCLASSOPTIONcaptionsoff
  \newpage
\fi

\end{document}